\documentclass[10pt,twocolumn,letterpaper]{article}

\usepackage{cvpr}
\usepackage{times}
\usepackage{epsfig}
\usepackage{graphicx}
\usepackage{amsmath}
\usepackage{amssymb}
\usepackage{booktabs}
\usepackage{subfig}
\usepackage{enumitem}
\makeatletter
\@namedef{ver@everyshi.sty}{}
\makeatother
\usepackage{tikz}
\usepackage{pgfplots}
\usepackage{bbm}
\usepackage[breaklinks=true,bookmarks=false]{hyperref}

\cvprfinalcopy % *** Uncomment this line for the final submission

\def\cvprPaperID{956} % *** Enter the CVPR Paper ID here

\begin{document}

%%%%%%%%% TITLE
\title{Unsupervised Open Domain Recognition by Semantic Discrepancy Minimization}

\author{\parbox{16cm}{\centering
		{\large \quad Junbao Zhuo$^{1,2}$ \quad\quad Shuhui Wang$^{1}$\thanks{Corresponding author.} \quad\quad Shuhao Cui$^{1,2}$
			\quad\quad Qingming Huang$^{1,2}$ }\\ 
		{\normalsize
			$^1$ Key Lab of Intell. Info. Process., Inst. of Comput. Tech., CAS, Beijing, 100190, China\\
			$^2$ University of Chinese Academy of Sciences, Beijing, 100049, China\\
		}
		{\tt\small junbao.zhuo@vipl.ict.ac.cn \quad \{wangshuhui,cuishuhao18s\}@ict.ac.cn \quad qmhuang@ucas.ac.cn 
		}
	}
}

\maketitle
\thispagestyle{empty}
\pagestyle{empty}
%%%%%%%%% ABSTRACT
\begin{abstract}
   We address the unsupervised open domain recognition (UODR) problem, where categories in labeled source domain $\mathcal{S}$ is only a subset of those in unlabeled target domain $\mathcal{T}$. The task is to correctly classify all samples in $\mathcal{T}$ including known and unknown categories. UODR is challenging due to the domain discrepancy, which becomes even harder to bridge when a large number of unknown categories exist in $\mathcal{T}$. Moreover, the classification rules propagated by graph CNN (GCN) may be distracted by unknown categories and lack generalization capability. 
   
   To measure the domain discrepancy for asymmetric label space between $\mathcal{S}$ and $\mathcal{T}$, we propose Semantic-Guided Matching Discrepancy~(SGMD), which first employs instance matching between $\mathcal{S}$ and $\mathcal{T}$, and then the discrepancy is measured by a weighted feature distance between matched instances. We further design a limited balance constraint to achieve a more balanced classification output on known and unknown categories. We develop Unsupervised Open Domain Transfer Network~(UODTN), which learns both the backbone classification network and GCN jointly by reducing the SGMD, enforcing the limited balance constraint and minimizing the classification loss on $\mathcal{S}$. UODTN better preserves the semantic structure and enforces the consistency between the learned domain invariant visual features and the semantic embeddings. Experimental results show superiority of our method on recognizing images of both known and unknown categories.
\end{abstract}

\vspace{-2ex}
\section{Introduction}
We study the unsupervised open domain recognition problem~(UODR) in this paper. In UODR, a labeled source domain $\mathcal{S}$ and unlabeled target domain $\mathcal{T}$ are given, where the categories in $\mathcal{S}$ is only a subset of those in $\mathcal{T}$. The task is to classify {\it all} samples in $\mathcal{T}$ including known  and unknown categories, which is undoubtedly a more challenging task but closer to the case in real-world applications compared to other related tasks in Domain Adaptation (DA)~\cite{oda,pda,pmd,dan,jan,dcor,wpda,ducda,tzeng2017adversarial,hoffman2017cycada,courty2017optimal,gong2012geodesic,pan2011domain} and Zero-Shot Learning (ZSL)~\cite{NIPS2009_3650,sharemodel,pst,udaz,ubias,gcn,adgpm,romera2015embarrassingly,zhang2017learning}.

\begin{table*}[]
	\centering
	\caption{The major differences among UODR, DA and ZSL problems.}
	\begin{tabular}{cccccc}
		\toprule & \begin{tabular}[c]{@{}c@{}}Domain \\ discrepancy\end{tabular} & \begin{tabular}[c]{@{}c@{}}Unknown \\ classes in $\mathcal{T}$\end{tabular} & \begin{tabular}[c]{@{}c@{}}Classify all \\samples in $\mathcal{T}$\end{tabular} & \begin{tabular}[c]{@{}c@{}}Use $\mathcal{T}$\\ for training\end{tabular}  & \begin{tabular}[c]{@{}c@{}}Asymmetric \\ label space\end{tabular} \\
		\midrule Unsupervised DA      &\checkmark              &$\times$                       &\checkmark                          &\checkmark                 &$\times$                 \\
		Partial Unsupervised DA       &\checkmark              &$\times$                       &\checkmark                          &\checkmark                 &\checkmark               \\
		Unsupervised open set  DA     &\checkmark              &\checkmark                     &$\times$                            &\checkmark                 &$\times$                 \\
		Generalized ZSL               &$\times$                &\checkmark                     &\checkmark                          &$\times$                   &--                       \\
		Transductive Generalized ZSL  &$\times$                &\checkmark                     &\checkmark                          &\checkmark                 &--                       \\
		UODR                          &\checkmark              &\checkmark                     &\checkmark                          &\checkmark                 &\checkmark               \\
		\bottomrule
	\end{tabular}
	\label{table:diff}
\end{table*}
The major differences among UODR and other related problems are summarized in Table \ref{table:diff}. Traditional unsupervised DA~\cite{pmd,dan,jan,dcor,ducda} is too strict to assume that $\mathcal{S}$ and $\mathcal{T}$ share the same categories.
Researchers begin to explore a more difficult setting that $\mathcal{S}$ and $\mathcal{T}$ do not share the same categories (asymmetric category space). In partial adversarial-DA~\cite{pda} and partial weighted-DA~\cite{wpda}, the authors addressed the problem that the category space of $\mathcal{T}$ is a subset of $\mathcal{S}$. However, the category space is still constrained in close set expanded by source domain categories. For the more difficult setting, i.e., the category space of $\mathcal{S}$ is a subset of $\mathcal{T}$ is rarely considered in DA field.
In open set DA~\cite{oda}, there are unknown categories both in $\mathcal{S}$ and $\mathcal{T}$, but the task is to classify only the samples of known categories in the target domain, while the samples of unknown categories are ignored. In contrast, there is no unknown categories in $\mathcal{S}$, and all samples of known and unknown categories in target domain are required to be classified in UODR.
UODR is also different from generalized ZSL~\cite{pst,udaz,sharemodel,ubias,gcn,adgpm}, since in generalized ZSL all the data are from the same domain and there is no presumed domain discrepancy between $\mathcal{S}$ ({\it i.e.}, training set in ZSL) and $\mathcal{T}$ ({\it i.e.}, testing set in ZSL).
Therefore, existing solutions can not be directly used to solve UODR problem due to its unique characteristics.

UODR is challenging due to the {\it semantic discrepancy} between $\mathcal{S}$ and $\mathcal{T}$, which can be explained from both feature distribution and semantic aspects.
First, there is large divergence on both content and distribution perspectives between $\mathcal{S}$ and $\mathcal{T}$, which is also referred to as domain discrepancy in existing DA studies~\cite{pmd,dan,jan,dcor,ducda}. The domain discrepancy is even harder to bridge if a large number of unknown categories are injected into $\mathcal{T}$. In this case, directly applying techniques used in DA, {\it e.g.}, MMD~\cite{dan} and DCORAL~\cite{dcor}, would lead to negative transfer. 
Second, it is hard to classify instances of unknown categories without labeled training data or any auxiliary attributes information~\cite{udaz,wang2012multi}. 
With the knowledge on the relationship among known and unknown categories, graph CNN (GCN)~\cite{gcn} can be used in UODR to propagate classification rules of known categories to unknown categories~\cite{zgcn,adgpm}. However, in generalized ZSL, there exists mode collapse that forces the prediction of unknown categories samples into the seen categories. Worse still, the propagated classification rules on unknown categories may lack generalization capability due to the domain discrepancy between $\mathcal{S}$ and $\mathcal{T}$.

The key idea to address UODR is minimizing the semantic divergence from both feature distribution and semantic aspects. Specifically, on unlabeled domain $\mathcal{T}$, there exists many unknown categories with similar image instances given a certain known category in $\mathcal{S}$. To reduce the distraction brought by unknown categories in $\mathcal{T}$, the domain-invariant feature learning is performed by reducing the domain discrepancy measured on data from the shared (known) categories of $\mathcal{S}$ and $\mathcal{T}$. We propose Semantic-Guided Matching Discrepancy~(SGMD), which first employs instance matching between $\mathcal{S}$ and $\mathcal{T}$ to produce coarsely matched pairs~\cite{pda}. The discrepancy is then measured by a weighted feature distances on these pairs, where the weight is the thresholded similarity of their target domain classifier responses.
The target domain classification output provides semantic level abstraction on a wide range of categories, and instance pair with the same category label are assumed to have similar classification outputs. Therefore, the weight reflects the degree of semantic consistency of each pair, and the weighted distance calculation further reduces the negative effect of noisy matching.

Similar as~\cite{zgcn,adgpm}, GCN is used to propagate the classification rules from known to unknown categories as the first step, where the category relation is described by WordNet. The propagated classification rules are then used to initialize the classification layer of backbone network. Based on the backbone classification network, to deal with semantic shift from known to unknown categories, we design a limited balance constraint to prevent target domain samples of unknown categories being classified into known categories, and better avoid strongly biased classifiers on unknown categories compared to the balance constraint proposed by~\cite{ubias}.

Putting the components together, we develop Unsupervised Open Domain Transfer Network~(UODTN), which learns the backbone classification network and GCN {\it jointly} by reducing the SGMD, achieving the limited balance, enforcing the semantic structure preserving via GCN, and minimizing the classification loss on $\mathcal{S}$.
Compared to multi-stage learning paradigms~\cite{zgcn,adgpm} that perform GCN-based classification model propagation and visual feature learning step-by-step, the joint classification network and GCN learning can better preserve the semantic structure and enforce the consistency between the learned domain invariant visual features and the semantic embeddings.
We construct two datasets for evaluating our method on UODR. Experimental results show the effectiveness of our method on recognizing images of both known and unknown categories in $\mathcal{T}$. We make our collected data and codes publicly available at https://github.com/junbaoZHUO/UODTN.

\section{Related Work}
\textbf{Deep unsupervised domain adaptation.}
Most of the deep unsupervised domain adaptation models are trained by combining classification loss on $\mathcal{S}$ with additional losses such as discrepancy reducing losses~\cite{dan,dcor,jan,pmd,Ghifary2015Domain}, adversarial discriminative losses~\cite{Ganin2016Domain,tzeng2017adversarial,volpi2018adversarial}, adversarial generative losses~\cite{Liu2016Coupled,bousmalis2017unsupervised,hoffman2017cycada} and reconstruction losses~\cite{Ghifary2016Deep}. We only review some  discrepancy-reducing-based methods that closely related to our method. A single linear kernel is applied to only one Fully-Connected~(FC) layer to minimize Maximum Mean Discrepancy (MMD) in DDC~\cite{Tzeng2014Deep}. The sum of MMDs defined between several FC layers, including the last classification layer, is considered in Deep Adaptation Network~(DAN)~\cite{dan}. In Joint Adaptation Networks~\cite{jan}, the joint distribution discrepancies of the multi-layer activations are considered rather than separate adaptations on marginal and conditional distributions which often require strong independence and/or smoothness assumptions on the factorized distributions. Instead of MMD, domain discrepancy is measured by the difference between the second-order statistics ({\it i.e.}, covariance)~\cite{dcor,ducda}. Domain discrepancy on both convolutional representation and the classification layer is explicitly considered in~\cite{ducda}. PMD~\cite{pmd} aims to approximate the first-order Wasserstein distance between two domains via minimum weight graph matching. These discrepancy-reducing-based methods can only handle the case that $\mathcal{S}$ and $\mathcal{T}$ share the same label space.

\textbf{Generalized ZSL.} Generalized ZSL drops the assumption that target domain contains only unknown categories~\cite{norouzi2013zero,kodirov2017semantic,sharemodel,ubias,udaz,pst,NIPS2013_5027}. Being the most related problem to UODR, transductive generalized ZSL~\cite{sharemodel,ubias,pst,udaz} is performed in a semi-supervised learning manner that both the labeled source data and the unlabeled target data are available, where there is no presumed domain discrepancy between $\mathcal{S}$ and $\mathcal{T}$. However, in UODR, there exists domain discrepancy between $\mathcal{S}$ and $\mathcal{T}$. Propagated Semantic Transfer (PST)~\cite{pst} exploits the manifold structure of novel classes by incorporating external knowledge, such as linguistic or expert specified information to conduct label propagation. Unsupervised Attribute Alignment (UAA)~\cite{udaz} associates cross-domain attributes by regularized sparse coding which enforces attributes shared by known and unknown categories to be similar. In~\cite{sharemodel}, a novel joint learning approach is proposed to learn the shared model space (SMS) for models such that the knowledge can be effectively transferred between classes using the attributes. Unbias ZSL~\cite{ubias} enforces a balanced classifier responses among known and unknown categories for unlabeled target data to learn an unbiased embedding space for ZSL.

\textbf{Object recognition via knowledge graph.} Salakhutdinov {\it et al.}~\cite{salakhutdinov2011learning} use WordNet to share the representations among different object classifiers so that objects with few training examples can borrow statistical strength from related objects. Deng {\it et al.}~\cite{deng2014large} apply the exclusion rules as a constraint and add object-attribute relations into the graph to train object classifiers for zero-shot applications. In contrast to these methods of using graph as constraints, a 6-layer deep GCN is constructed to directly generate novel object classifiers in~\cite{zgcn}. In~\cite{adgpm}, the authors argue that too many layers of GCN results in over-smooth classifier and propose to train a single layer GCN. Furthermore, in \cite{adgpm}, a more dense graph structure is utilized and fine-tune the feature space to adapt to the generated semantic embedding space.

\section{Method}
\begin{figure*}
	\centering
	\includegraphics[width=.97\textwidth]{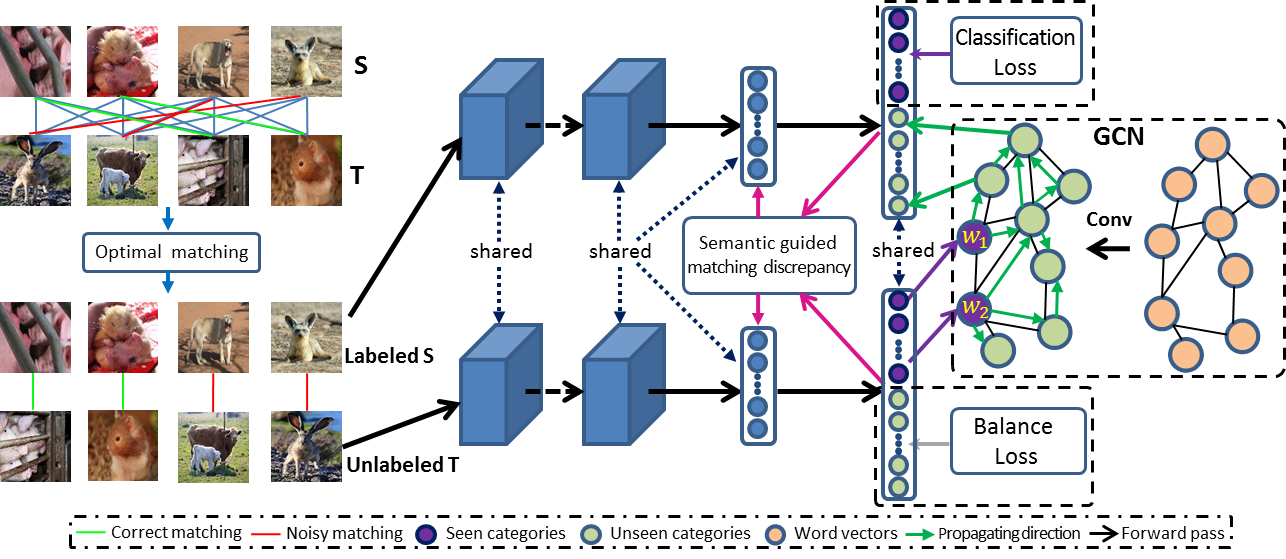}
	\caption{The proposed UODTN framework for UODR problem. It consists of a two-stream Siamese network representing the source and target models where weights of all layers are shared and a GCN for propagating classification rules of known categories in source domain to unknown categories in target domain. The Siamese network and GCN are jointly trained in an end-to-end manner. The proposed semantic-guided matching discrepancy is estimated on the features extracted from source and target domain. By reducing the proposed discrepancy, UODTN is able to propagate more suitable source classifiers to unknown categories in target domain as the source classifiers are based on domain-invariant features.} %To prevent the unknown classifier biasing toward known classifier}
	\label{fig:framework}
	\vspace{-2ex}
\end{figure*}
\subsection{Common Notations}
Some common notations used in this paper are introduced here. Suppose that there are $N_S$ source-domain training examples $\mathcal{D}_S=\{z_i^s \}_{i=1}^{N_S}$ with labels $\mathcal{L}_S=\{y_i \}_{i=1}^{N_S},~y_i\in \{1,2,\ldots,L_S\}$, and $N_T$ unlabeled target-domain examples $\mathcal{D}_T=\{z_j^t \}_{j=1}^{N_T}$ , where their labels $\mathcal{L}_T=\{y_j \}_{j=1}^{N_T},~y_j\in \{1,2,\ldots,L_T\}$ are not available and $L_S<L_T$. That is, there are $L_T-L_S$ unknown categories in target domain. $z_i^s$ and $z_j^t$ are the raw images from source and target domains respectively. Let $\phi(\cdot)$ be the feature extractor and let $\psi_S(\cdot)$ and $\psi_T(\cdot)$ denote the classifier pretrained on $\mathcal{S}$ and classifier for target domain $\mathcal{T}$. 

\subsection{Framework}

As shown in Figure~\ref{fig:framework}, our Unsupervised Open Domain Transfer Network (UODTN) contains a backbone classification network with classifier layer for all $L_T$ categories in target domain and a GCN that maintains the relationships among all $L_T$ categories. We first use GCN to generate the semantics embeddings of unknown categories in target domain and then initialize the classifier layer of backbone classification network by these semantic embeddings. Based on the initialized backbone classification network, we further reduce the proposed semantic-guided matching discrepancy, enforce the proposed limited balance constraint and integrate GCN to minimize the semantic discrepancy in UODR problem. The backbone classification network and GCN are jointly trained in an end-to-end manner with GCN aiming at preserving semantic structure encoded in word vectors and knowledge graph. The details are illustrated as follows.

\subsection{Generating unknown class semantic embeddings} \label{initial}
With the auxiliary information encoded in word vectors and knowledge graph for unknown categories, we can generate the unknown class semantic embeddings via GCN. We first construct a graph with $N$ nodes where each node is a $C$-dimensional vector presenting a distinct concept/class. In order to propagate the semantic embeddings of known categories to unknown categories, additional nodes are required for constructing full path from known categories to unknown categories. Each node is initialized with word vector of the class name. The relationships among the classes in the knowledge graph, say, WordNet, are encoded in form of a symmetric adjacency matrix $A\in R^{N\times N}$, which also includes self-loops. We propagate such relationship as performing convolution on the graph
$$O=\sigma(D^{-1}AX\Theta)\eqno{(1)}$$
where $X\in R^{N\times C}$ is composed of $N$ word vectors and $\Theta \in R^{C\times F}$ denotes the trainable weights. $\sigma(\cdot)$ denotes a nonlinear activation function. $D\in R^{N\times N}$ is a degree matrix where $D_{ii}=\sum_jA_{ij}$. By training the GCN to predict the classifier weights of known classes, the GCN simultaneously generates the classifier weights of unknown classes while preserves the semantic relationship exhibited in word vectors and knowledge graph. The loss is
$$\mathcal{L}_{init}=\frac{1}{2M} \sum_{i=1}^{L_S}\sum_{j=1}^M(O_{i,j}-W_{i,j})^2 \eqno{(2)}$$
where $W\in R^{{L_S}\times M}$ denotes the classifier weights obtained by extracting the weights of $\psi_S(\cdot)$, the classifier pretrained on source domain. We replace the original classifier of pretrained ResNet-50 with the generated classifiers to form a classification network for source and target domain.

\subsection{Semantic-guided matching discrepancy}
In real world scenario, there always exists domain discrepancy between manually collected labeled data (source domain) and practical data (target domain). Such domain discrepancy leads to performance degradation on target domain and more severely, makes GCN propagate biased semantic embeddings to unknown categories. Therefore, it is urgent to reduce the domain discrepancy. However, it is difficult to measure the domain discrepancy in UODR problem since there are many unknown categories samples. Existing domain discrepancy measurements such as MMD~\cite{dan,jan} and difference between correlation~\cite{dcor,ducda}, assume that the source and target domain share same categories, which can not handle asymmetric label space for UODR.

We propose semantic-guided matching discrepancy to estimate the domain discrepancy. We extract the features of all instances from source and target domain and construct a bipartite graph between the two domains. The weights of the bipartite graph are pairwise distance of all pairs. In this work, we use $L_1$ distance while other distance metrics can also be used. By solving minimum weight matching problem via the Hungarian algorithm, we obtain coarse and noisy matched instance pairs~(pairs linked with red line in the left part of Figure~\ref{fig:framework}) between source and target domain. Directly reducing the discrepancy measured from noisy matched instances pair will inevitably lead to negative transfer. Hence, we propose to utilize the semantic consistency of matched pairs to filter such noisy matched pairs. Precisely, given matched source and target instances $z_i^s$ and $z_i^t$, we extract their features as $f_i^s=\phi(z_i^s)$ and $f_i^t=\phi(z_i^t)$, and calculate their classifier responses $p_i^s=\psi_T(f_i^s)$ and $p_i^t=\psi_T(f_i^t)$ respectively, the semantic-guided matching discrepancy is
$$\mathcal{L}_{d}=\sum_id(f_i^s,f_i^t) * \mathbbm{1}(\langle p_i^s,p_i^t\rangle > \tau) \eqno{(3)}$$
where $d(f_i^s,f_i^t)$ is the distance metrics which can be $L_2$ distance, the discrepancy metric encoded in domain discriminator when using adversarial training, etc. $\langle \cdot,\cdot \rangle$ denotes inner product. $\mathbbm{1}$ is indicator function and $\tau$ is a given threshold. The similarity $\langle p_i^s,p_i^t\rangle$ reveals the degree of semantic consistency of each pair since samples of the same classes are assumed to have similar classification responses.

\subsection{Limited balance constraint}

To prevent target domain samples of unknown categories being classified into known categories, it is straightforward to add a balance constraint to classifier responses for target domain instances. The vanilla balance constraint~\cite{ubias} is calculated as:
$$\mathcal{L}_b=-log\sum_{j=L_S+1}^{L_T}p_j^t \eqno{(4)}$$
However, such balance constraint may grow into unexpected large value since there is no label for target domain, which will result in biased classifiers of unknown categories. To prevent the classifier response of unknown categories growing abnormally, we propose limited balance constraint:
$$\mathcal{L}_{lb}=R_t + \frac{w^2}{R_t} \eqno{(5)}$$
where $R_t=\sum_{j=L_S+1}^{L_T}p_j^t$ and $w$ is a manually set constant that control the ratio of classification response of unknown categories over all categories. Such constraint enforces the ratio of classification response of unknown categories over all categories to lie in an appropriate range. Ideally, $w$ can be set according to the prior of the proportion of unknown classes over all categories.

\subsection{Semantic structure preserving via GCN}
The semantic structure among categories exhibited in word vectors and knowledge graph can not be well preserved via reducing semantic-guided matching discrepancy and enforcing limited balance constraint. To preserve such relationship, we integrate GCN into our training, resulting in an end-to-end framework. Different from subsection \ref{initial}, semantic embeddings of all categories in target domain are considered in the loss term:

$$\mathcal{L}_{gcn}=\frac{1}{2M} \sum_{i=1}^{L_T}\sum_{j=1}^M(O_{i,j}-\widehat W_{i,j})^2 \eqno{(6)}$$
where $\widehat W\in R^{{L_T}\times M}$ denotes the classifier weights obtained by extracting the weights of $\psi_T(\cdot)$, the classification layer for all categories in target domain. Unlike the method proposed in~\cite{adgpm}, which fixes the classifier learnt from GCN and fine-tune the features, the classifier in our model can be well adapted to data while the semantic relationship of all categories is still maintained via GCN.
\subsection{Joint training}
After initializing the classifier layer of UODTN via trained GCN in subsection~\ref{initial}, we utilize all proposed techniques to train UODTN in an end-to-end manner. The total loss is
$$\mathcal{L}=\mathcal{L}_{cls}+\lambda_d\mathcal{L}_d+\lambda_{b}\mathcal{L}_{lb}+\lambda_{g}\mathcal{L}_{gcn} \eqno{(7)}$$
where $\mathcal{L}_{cls}$ is classification loss on labeled source domain. $\lambda_d$, $\lambda_b$ and $\lambda_g$ are weights for semantic-guided matching discrepancy minimizing loss, limited balance constraint and structure preserving loss of GCN. Specifically, minimizing semantic-guided matching discrepancy provides domain-invariant features for classifiers of known and unknown categories. Further, the classifiers of known categories receive both the supervision of classification loss and regularization of GCN. On the other hand, the classifiers of unknown categories are trained with guidance from limited balance constraint and GCN. Joint training is unhindered to achieve better trade-off of classification accuracy between known and unknown categories in target domain. Minimizing sematic guided matching discrepancy actually propagates semantic information from feature perspective while GCN propagates semantic embeddings from semantic perspective. The UODR problem is actually an ill-conditioned problem where limited balance constraint prevents ill solutions of UODTN during the training progress. %Therefore, joint training of backbone classification network and GCN can

\vspace{-2ex}
\section{Experiment}

\subsection{Datasets} 
\begin{figure}
	\centering
	\subfloat{
		\begin{minipage}{.113\textwidth}\includegraphics[width=1\textwidth]{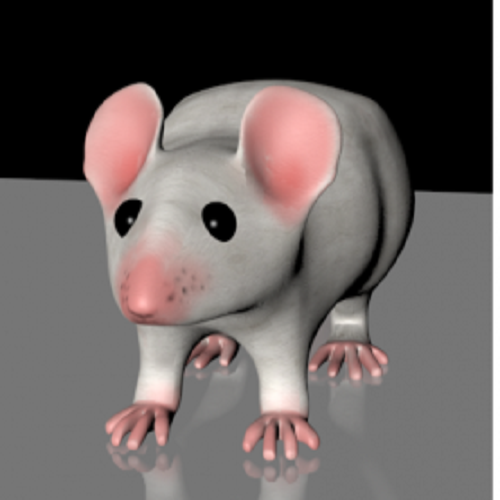}\end{minipage}%\hspace{-4pt}
	}
	\subfloat{
		\begin{minipage}{.113\textwidth}\includegraphics[width=1\textwidth]{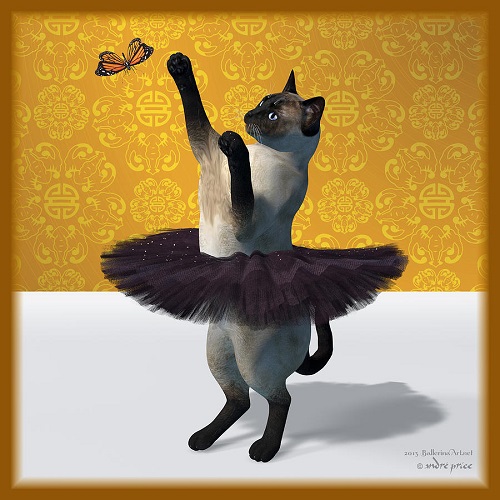}\end{minipage}%\hspace{-4pt}
	}
	\subfloat{
		\begin{minipage}{.113\textwidth}\includegraphics[width=1\textwidth]{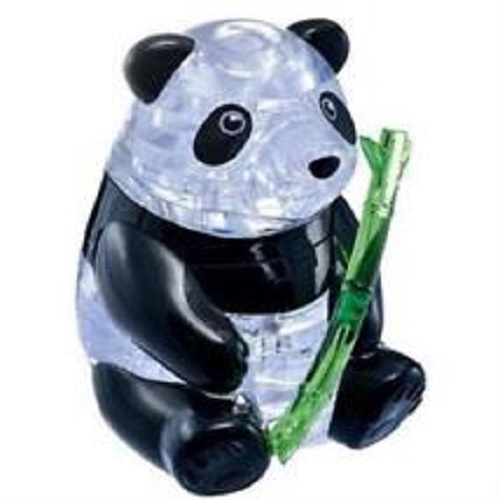}\end{minipage}%\hspace{-4pt}
	}
	\subfloat{
		\begin{minipage}{.113\textwidth}\includegraphics[width=1\textwidth]{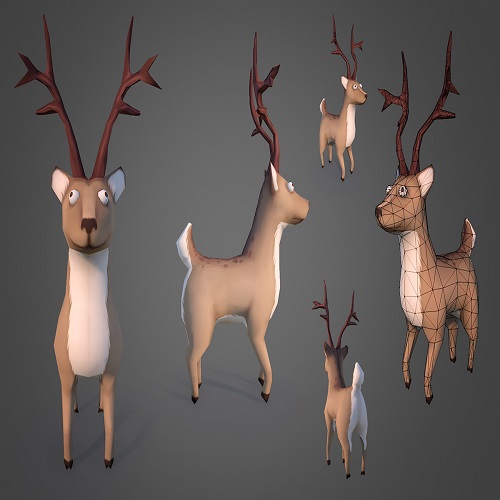}\end{minipage}%\hspace{-4pt}
	}
	\vspace{-10pt}
	\\
	\subfloat{
		\begin{minipage}{.113\textwidth}\includegraphics[width=1\textwidth]{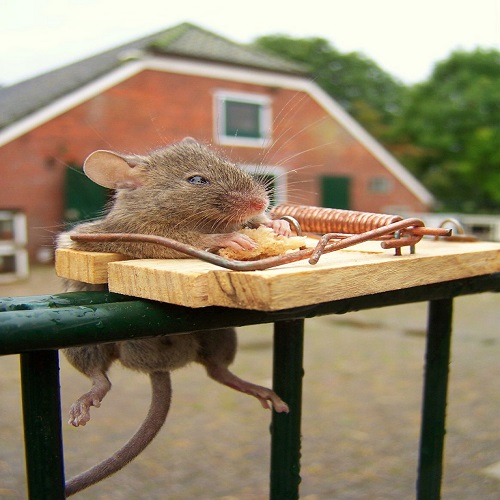}\end{minipage}%\hspace{-4pt}
	}
	\subfloat{
		\begin{minipage}{.113\textwidth}\includegraphics[width=1\textwidth]{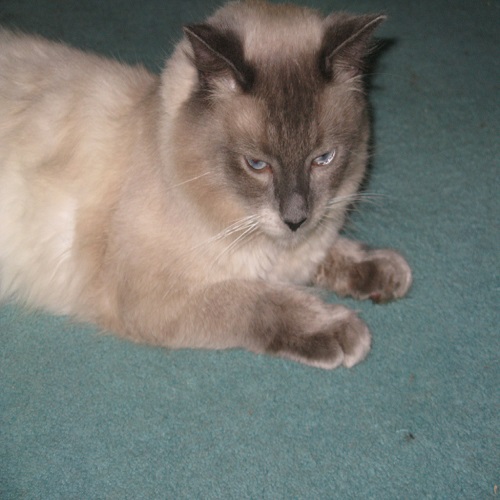}\end{minipage}%\hspace{-4pt}
	}
	\subfloat{
		\begin{minipage}{.113\textwidth}\includegraphics[width=1\textwidth]{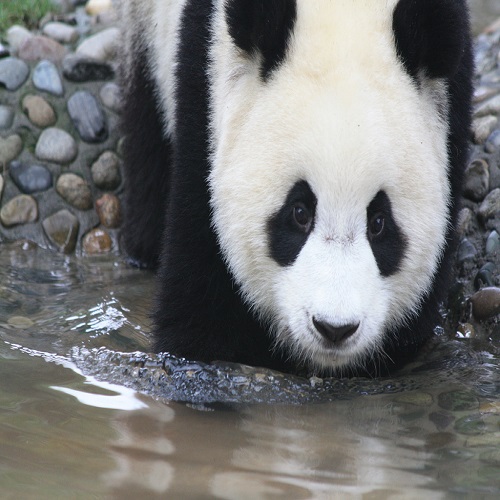}\end{minipage}%\hspace{-4pt}
	}
	\subfloat{
		\begin{minipage}{.113\textwidth}\includegraphics[width=1\textwidth]{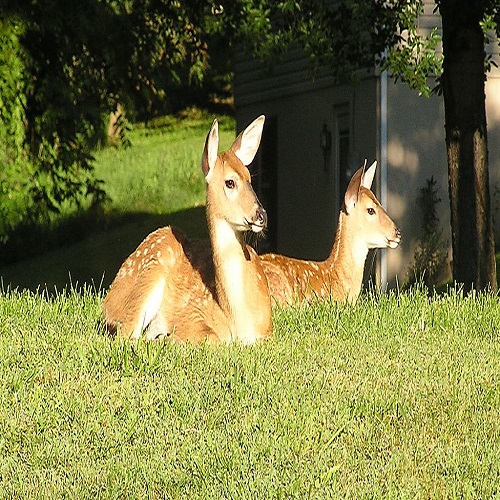}\end{minipage}%\hspace{-4pt}
	}
	%%%%}
	%\setlength{\abovecaptionskip}{4pt}
	\caption{The first and second rows show the samples from source and target domain of I2AwA respectively and images of the same column belong to the same categories. The target samples are taken from natural scene while source data that are collected from Internet, containing 3D model images of animals, which appear to be discrepant from target domain on both content and distribution perspectives.}
	\label{fig:sample}
	\vspace{-1ex}
\end{figure}

We evaluate our method on two datasets: a small-scale dataset I2AwA and a large-scale dataset I2WebV. The target domain of I2AwA is AwA2\cite{aw2} which is a replacement of the original AwA dataset for zero-shot learning. It consists of 50 animal classes, with a total of 37,322 images and an average of 746 images per class. We use the proposed split in~\cite{aw2} in which 40 classes are regarded as known categories and the rest 10 classes as unknown categories. We collect a source domain dataset with 40 known categories via Google image searching engine. We manually remove the noisy images resulting 2,970 images in total. There exists domain discrepancy between source and target domain as shown in Figure\ref{fig:sample}. As for I2WebV, its source domain is ILSVRC-2012 with 1,000 classes which consists of 1,279,847 images totally. The target domain of I2WebV is the validation set of WebVision~\cite{webv} with 5,000 classes, which is composed of 294,009 images. I2WebV is a very challenging dataset as there is large domain discrepancy between two domains and large number of unknown categories in target domain, some of which are very different from 1,000 known categories. The knowledge database we use for both I2AwA and I2WebV is WordNet~\cite{wordnet}.

\subsection{Evaluation metrics} We perform classification on the whole target domain similar to generalized zero-shot learning and report the Top 1 Accuracies of known categories, unknown categories and all categories on target domain for better understanding the knowledge transfer process.

\subsection{Baselines} we compare our method with several baselines: \textbf{zGCN}~\cite{zgcn}, two variants including \textbf{dGCN} and \textbf{adGCN} proposed in~\cite{adgpm}, \textbf{bGCN} and \textbf{pmd-bGCN}. \textbf{zGCN} is built upon graph which utilizes both word vectors and the categorical relationships encoded in WordNet to predict the classifiers of unknown categories. Following \textbf{zGCN}, the authors in~\cite{adgpm} utilize a more dense graph structure (\textbf{dGCN}) and assign different weights for additional edges (\textbf{adGCN}). We also construct \textbf{bGCN}, GCN with original balance constraint proposed in state-of-the-art transductive zero-shot learning methods~\cite{ubias}. Furthermore, on the basic of \textbf{bGCN}, we implement another variant of GCN, \textbf{pmd-bGCN}, which further reduces the population matching discrepancy~\cite{pmd}, a state-of-the-art domain discrepancy measurement which shows superiority over MMD.

\begin{table}
	\caption{Top1 Accuracies on I2AwA.}
	\vspace{-2.5ex}
	\begin{center}
		\scalebox{1.1}{
			\begin{tabular}{cccc}
				\toprule                     &Known        &Unknown         &All          \\
				\midrule zGCN~\cite{zgcn}    &77.2         &21.0            &65.0         \\
				dGCN~\cite{adgpm}            &78.2         &11.6            &64.0         \\
				adGCN~\cite{adgpm}           &77.3         &15.0            &64.1         \\
				bGCN~\cite{ubias}            &84.6         &28.0            &72.6         \\	
				pmd-bGCN~\cite{pmd}          &\textbf{84.7}         &27.1            &72.5         \\		
				UODTN                        &\textbf{84.7}&\textbf{31.7}   &\textbf{73.5}\\
				\bottomrule
			\end{tabular}
		}
	\end{center}
	\label{table:I2AwA}
	\vspace{-4.5ex}
\end{table}

\begin{table}
	\caption{Top1 Accuracies on I2WebV.}
	\vspace{-2ex}
	\begin{center}
		\scalebox{1.05}{
			\begin{tabular}{cccc}
				%\toprule         & Acc on Seen& Acc on Unseen & Acc on All\\
				\toprule                      &Known        &Unknown         &All          \\			
				\midrule zGCN~\cite{zgcn}     &43.8         &2.2             &11.1         \\
				dGCN~\cite{adgpm}             &45.2         &2.0             &11.3         \\
				adGCN~\cite{adgpm}            &45.8         &2.2             &11.6         \\
				bGCN~\cite{ubias}             &47.4         &2.2             &12.0         \\	
				pmd-bGCN~\cite{pmd}           &47.2         &2.2             &11.9         \\
				UODTN (Ukn.)                  &51.9         &\textbf{3.2}    &13.8         \\
				UODTN (Avg.)                  &\textbf{57.3}&2.4             &\textbf{14.2}\\
				\bottomrule
			\end{tabular}
		}
	\end{center}
	\label{table:I2WebV}
	\vspace{-4ex}
\end{table}

\subsection{Implementation details} We construct two distinct graphs based on WordNet~\cite{wordnet} for I2AwA and I2WebV respectively. The graph nodes include all categories of target domain and their children and ancestors. Precisely, the number of nodes for graphs of I2AwA and I2WebV are 255 and 7,460. The word vectors for all categories are extracted via GloVe text model~\cite{GloVe} which is trained on Wikipedia. Word vectors for nodes in graph are set as inputs of GCN. We use ResNet-50~\cite{resnet} pretrained on ILSVRC-2012 as basic model where the last fully connected layer, i.e., the classification layer is regarded as the target that GCN tends to predict. We train the GCN with word vectors as inputs and classifier of pretrained ResNet-50 as target to obtain the initial classifiers of target domain in I2WebV. As for I2AwA, the supervison information for training GCN is classifiers finetuned on the source domain of I2AwA. These initial classifiers are then concatenated into feature extractor of pretrained ResNet-50~(with original classifier layer removed) to form a backbone classification network for source and target domain. We fix some beginning convolutional layers of ResNet-50 to accelerate the training process. The global average pooling responses before classification layer are thought as features and based on these features we construct a bipartite graph with each sub-graph representing source and target domain. We use Hungarian algorithm to get minimum weight matched pairs for estimating population matching discrepancy~\cite{pmd} and our proposed semantic-guided matching discrepancy. Specifically, we use the discrepancy metric encoded in domain discriminator as distance metric in Eqn. (3). It is difficult to get minimum weight matched pairs for bipartite graph based on large scale datasets. We simply apply \textbf{divide and conquer} strategy to handle this issue. Take I2AwA as an example, we randomly divide source/target domain into 5 folds, respectively. Then we construct 5 bipartite graphs for each fold pair and use Hungarian algorithm to get minimum weight matched pairs for 5 bipartite graphs. All of our experiments are implemented with Pytorch~\cite{torch}. More details can be seen in our released codes.

\subsection{Results and discussion} The classification results on I2AwA and I2WebV are shown in Table~\ref{table:I2AwA} and Table~\ref{table:I2WebV}. As shown in Table~\ref{table:I2AwA} and Table~\ref{table:I2WebV}, our method UODTN outperforms all the baselines by considerable margins, achieving 3.7\% and 0.9\% improvements on unknown classes and all classes on I2AwA. For a more challenging dataset I2WebV, we implement two variants of UODTN with different $\lambda_d$, $\lambda_b$ and $\lambda_g$ according to different trade-off between known and unknown categories. Precisely, aiming at achieving higher average performance, UODTN (Avg.) shows 9.9\%, 0.2\% and 2.2\% improvements on known classes, unknown classes and all classes compared to bGCN. On the contrary, UODTN (Ukn.) that pays more attention to unknown categories, achieves remarkable improvement on unknown categories by 1.0\% while the overall top 1 accuracy is still higher that bGAN. Noting that WebVision contains 4,000 unknown categories, 1.0\% improvement is a great progress without any labels of unknown categories available. 
We also obtain the following observations: (1) zGCN, dGCN and adGCN obtained from labeled source domain and knowledge graph can not fit target data well, as there is severe classification confusion between known and unknown categories. UODTN and bGCN show improvement over zGCN, dGCN and adGCN indicating that fitting target domain data leads to better generalization of networks. However, comparing bGCN with UODTN, we can see that merely introducing a balance constraint on classifier responses is insufficient as there exists domain discrepancy between source and target domain. Such domain discrepancy results in suboptimal classifiers cause distracted semantic embeddings when being propagated to unknown categories in target domain. (2) Merely reducing the domain discrepancy estimated by traditional methods leads to negative transfer as revealed by comparison between bGCN and pmd-bGCN. Note that we assign a very small weight to population matching discrepancy reducing term for optimal results of pmd-bGCN. On the contrary, by reducing our proposed semantic-guided matching discrepancy, such negative transfer can be avoided and more domain-invariant features are learned by UODTN, which is illustrated in \ref{ablation}.

\begin{table}
	\caption{Ablation study on I2AwA.}
	\vspace{-2.5ex}
	\begin{center}
		\scalebox{0.96}{
			\begin{tabular}{cccc}
				\toprule                 &Known        &Unknown         &All          \\
				\midrule zGCN~\cite{zgcn}&77.2         &21.0            &65.0         \\
				UODTN (lb)               &83.9         &\textbf{32.5}            &73.0         \\
				UODTN (lb+sgmd)          &84.6         &31.0            &73.3         \\
				UODTN (lb+sgmd+gcn)      &\textbf{84.7}         &31.7            &\textbf{73.5}         \\
				\bottomrule
			\end{tabular}
		}
	\end{center}
	\label{table:ablation}
	\vspace{-6ex}
\end{table}

\begin{figure}
	\centering
	\includegraphics[width=.48\textwidth]{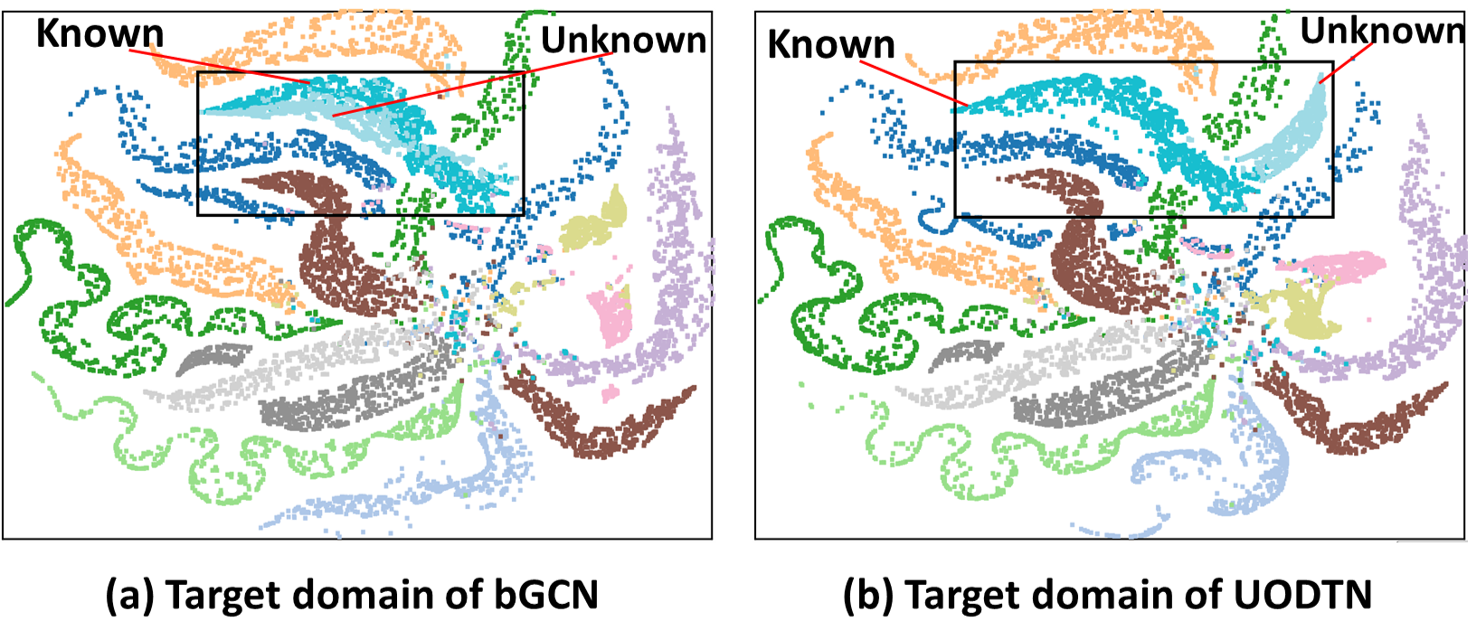}
	\caption{Visualizations of features learned by UODTN and bGCN in target domain of I2AwA. From the black boxes areas, we can see that samples of an unknown category are mixed with a known category for bGCN, while these two categories are well separated by UODTN. This validates that semantic discrepancy is alleviated by UODTN.}
	\vspace{-2ex}
	\label{fig:vis}
\end{figure}

\begin{figure}
	\centering
	\includegraphics[width=.48\textwidth]{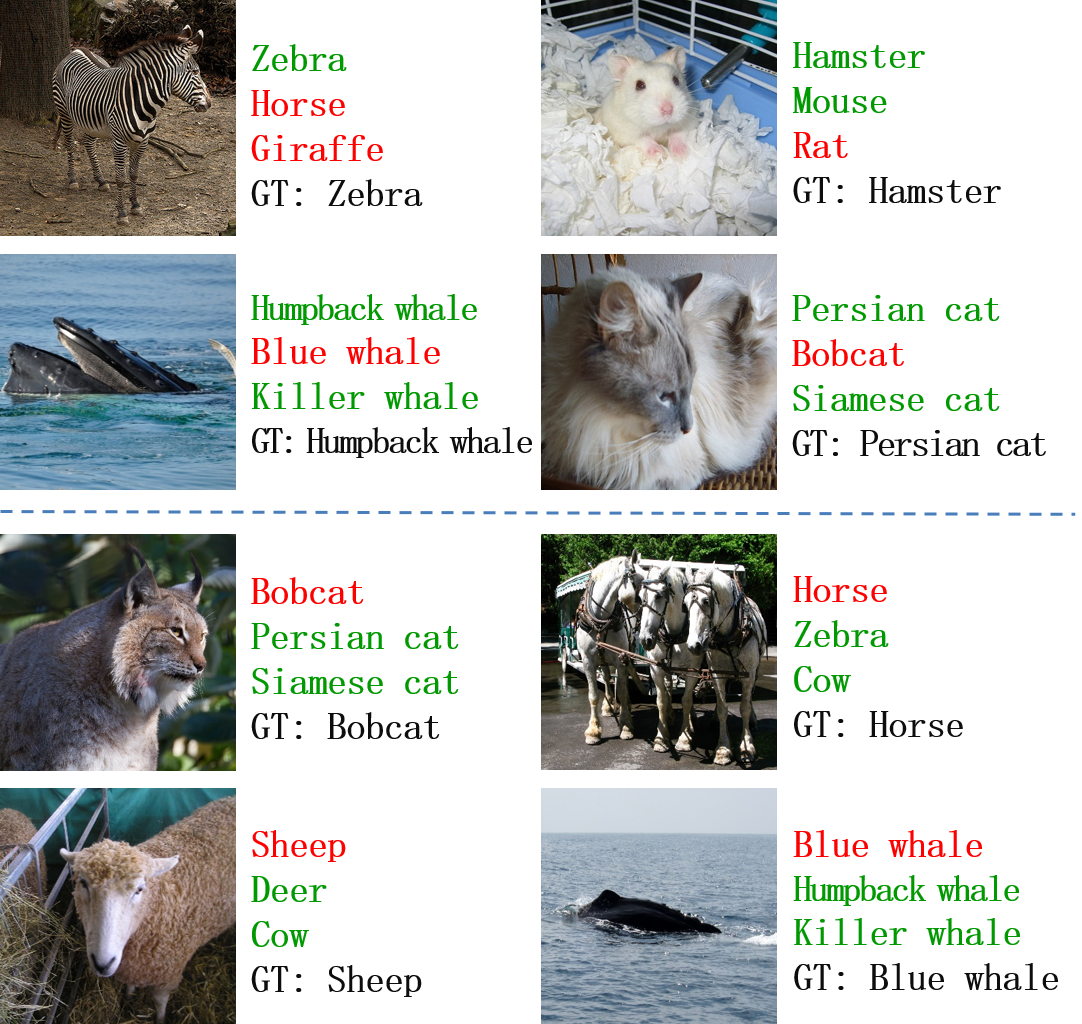}
	\caption{The top three classifier responses of UODTN with various target inputs. Green/red means that the category is known/unknown and GT means ground-truth. The first and second rows are examples of known categories and the rest are drawn from unknown categories. We can see that except ground-truth, UODTN assigns considerable weight on related unknown/known categories for known/unknown categories samples, indicating that the transferring mechanism of UODTN is effective. That is, the knowledge of labeled source domain, word vectors and WordNet can be transferred to unknown categories in a reasonable way.}
	\label{fig:case}
	\vspace{-2ex}
\end{figure}
\vspace{-1ex}
\subsection{Ablation Study}\label{ablation}
To go deeper with the efficacy of semantic-guided matching discrepancy, limited balance constraint and joint training of GCN, we conduct ablation study on I2AwA by evaluating several models (Table~\ref{table:ablation}): (1) zGCN, without adding any proposed techniques in UODTN; (2) UODTN (lb), which includes only limited balance constraint; (3) UODTN (lb+sgmd), which further contains semantic-guided matching discrepancy reducing module; (4) UODTN (lb+sgmd+gcn), which is the full model with limited balance constraint, semantic-guided matching discrepancy reducing module and joint training of GCN.
We can see that UODTN (lb) outperforms zGCN~\cite{zgcn} by a large margin since limited balance constraint can prevent the classifier activations on known categories growing abnormally. By the way, from Table~\ref{table:I2AwA} and~\ref{table:ablation}, we can see that UODTN (lb) outperforms bGCN which shows the superiority of limited balance constraint over original balance constraint~\cite{ubias}. Further, we can observe that UODTN (lb+sgmd) improves the performance by 0.1\% compared with UODTN (lb), which validates that reducing semantic-guided matching discrepancy can not only avoid negative transfer but also boost the domain invariance of learned features. By further integrating GCN for joint training, UODTN (lb+sgmd+gcn) gains improvement over UODTN (lb+sgmd). It is rational as the relationship among all known and unknown categories is essential for transferring effective semantic embeddings for unlabeled unknown categories. Joint training with GCN progressively maintains the semantic structure encoded in word vector and knowledge graph to guarantee the boost of UODTN.

\subsection{Traditional domain adaptation}
We conduct experiments for traditional domain adaptation to validate that semantic-guided matching discrepancy (SGMD) is capable of dealing DA. We simply adopt $L_2$ distance for Eqn. (3) here. The source domain is ImageNet and the target domain is a subset of Webvision that shares 1,000 categories with ImageNet for I2WebV. From the first row in Table~\ref{table:DA}, we can see that SGMD is slightly better than PMD and to MMD, demonstrating that weighted mechanism is helpful for DA. Note that the matching is fixed, so PMD is poor than MMD. However, our SGMD is still better than MMD which validates the effectiveness of weighted mechanism. Domain adaptation results on I2AwA are shown in the second row in Table~\ref{table:DA}. The discrepancy between source and target domain of I2AwA is large and the size of source domain is small. Besides, the categories in AwA2 are similar so that domain adaptation on I2AwA is very challenging. With fixed matching, SGMD outperforms MMD and PMD significantly which validates the superiority of SGMD.

\begin{table}
	\caption{Results for domain adaptation on I2WebV (the first row) and I2AwA (the second row).}
	\vspace{-3ex}
	\begin{center}
		\scalebox{1.0}{
			\begin{tabular}{ccccc}%c}
				\toprule              &         ResNet &MMD   &PMD   &SGMD      \\%&UODTN        \\
				\midrule   I2WebV (1K)&         67.7   &68.0  &67.9  &\textbf{68.1}      \\%&        \\	
				\midrule
				\midrule    I2AwA (40)&         84.0   &84.2  &84.4  &\textbf{85.1}      \\%&        \\
				\bottomrule
			\end{tabular}
		}
	\end{center}
	\label{table:DA}
	\vspace{-5ex}
\end{table}

\subsection{Visualization}
We visualize the t-SNE embeddings of the images of target domain with features extracted from best competitor bGCN and our model UODTN on I2AwA in Figure~\ref{fig:vis}. We only visualize 15 known categories and 3 unknown categories for the sake of visualization quality and clarity. These known categories include the categories that are related to 3 unknown for better understanding the influence between known and unknown categories. From Figure~\ref{fig:vis} (a), we can see that in the black box area, the samples of unknown category are mixed with those of known category for bGCN. On the contrary, in Figure~\ref{fig:vis} (b), the two categories are well separated by UODTN which qualitatively verifies the effectiveness of semantic-guided matching discrepancy, limited balance constraint and joint training of GCN in UODTN.% learns more discriminative features than bGCN.
%\vspace{-1ex}
\subsection{Illustrative examples}
We show some qualitative results of UODTN in Figure~\ref{fig:case}. We observe that UODTN effectively transfers the semantic embeddings of source domain to unknown categories in target domain. This property mainly depends on joint training with GCN to preserve the semantic relationships between known and unknown categories while improving the discrimination ability of classifier. Figure~\ref{fig:case} provides some correct classification results of UODTN. For all instances, except the true categories that the instances belong to, the classifiers of correctly related unknown/known categories are also activated with large confidence. This indicates that UODTN can effectively transfer the knowledge from both labeled source domain, word vectors and knowledge graph. More illustrative examples including incorrect results can be seen in supplementary material.

\vspace{-1ex}
\section{Conclusion}
We explore unsupervised open domain recognition problem, where an unlabeled target domain $\mathcal{T}$ and a discrepant labeled source domain $\mathcal{S}$ that only covers a subset of categories of target domain are given, and the goal is to classify all instances of target domain. UODR is more challenging due to the semantic discrepancy between $\mathcal{S}$ and $\mathcal{T}$, which exhibits large divergence on both content and distribution perspectives between $\mathcal{S}$ and $\mathcal{T}$ and semantic shift from known to unknown categories between the two domains. 
We develop Unsupervised Open Domain Transfer Network (UODTN) , which learns the backbone classification network and GCN {\it jointly} by reducing the SGMD, achieving the limited balance, enforcing the semantic structure preserving via GCN, and minimizing the classification loss on $\mathcal{S}$. 
We collect two datasets for UODR problem and extensive experiments validate the effectiveness of UODTN. 
In future work, discriminating known and unknown categories to alleviate the semantic shift in OUDR problem also worths studying, since it is a non-trivial task as there is function to distinguish known and unknown categories. 
\section{Acknowledgement}
This work was supported in part by National Natural Science Foundation of China: 61672497, 61620106009, U1636214 and 61836002, in part by National Basic Research Program of China (973 Program): 2015CB351800, and in part by Key Research Program of Frontier Sciences of CAS: QYZDJ-SSW-SYS013.

%\newpage
{\small
	\bibliographystyle{ieee}
	\bibliography{egbib}
}

\end{document}